\documentclass[10pt,letterpaper]{article}

\usepackage{cogsci}

\cogscifinalcopy % Uncomment this line for the final submission 

\usepackage{graphicx}
\usepackage{pslatex}
\usepackage{apacite}
\usepackage{float} % Roger Levy added this and changed figure/table
\usepackage{natbib}
\usepackage{hyperref}

\usepackage{algpseudocode}
\usepackage{algorithm}
\usepackage[dvipsnames]{xcolor}
\usepackage{booktabs}
\usepackage{array}
\usepackage{tabularx}
\usepackage{linguex}
\usepackage{multirow}
\usepackage{amsmath}
\usepackage{amssymb}
\usepackage{listings}
\usepackage{makecell}
\usepackage{microtype}      % microtypography
\usepackage{xcolor}         % colors
\usepackage{graphicx}       % figures (add this?)
\usepackage{tikz}
\usepackage[many]{tcolorbox}
\usepackage{textcomp}
\usepackage{multirow}
\usepackage{wrapfig}
\usepackage{caption}
\usepackage{soul}
\usepackage{adjustbox}
\usepackage{enumitem}
\usepackage{caption}
\usepackage{tabularx}
\definecolor{CB_gray}{gray}{0.5}

\tcbset{prompt/.style={
    enhanced,
    size=fbox,
    boxrule=3pt,
    arc=2mm,
    auto outer arc,
    left=10pt,
    right=10pt,
    top=10pt,
    bottom=10pt,
    fontupper=\fontfamily{cmtt}\selectfont, 
    colback=CB_gray!10,
    colframe=CB_gray!25,
    coltitle=CB_gray!100, 
}}

\tcbset{vignette/.style={
    enhanced,
    size=fbox,
    boxrule=3pt,
    arc=2mm,
    auto outer arc,
    left=10pt,
    right=10pt,
    top=10pt,
    bottom=10pt,
    fontupper=\fontfamily{Avenir}\selectfont, 
    colback=CB_gray!10,
    colframe=CB_gray!25,
    coltitle=CB_gray!100, 
}}

\definecolor{codegreen}{rgb}{0,0.6,0}
\definecolor{codegray}{rgb}{0.5,0.5,0.5}
\definecolor{codepurple}{rgb}{0.58,0,0.82}
\definecolor{backcolour}{rgb}{0.95,0.95,0.92}

\lstdefinestyle{mystyle}{
  commentstyle=\color{codegreen},
  keywordstyle=\color{magenta},
  numberstyle=\tiny\color{codegray},
  basicstyle=\ttfamily\footnotesize,
  stringstyle=\color{codepurple},
  breakatwhitespace=false,         
  breaklines=true,                 
  captionpos=b,                    
  keepspaces=true,                 
  numbers=left,                    
  numbersep=5pt,                  
  showspaces=false,                
  showstringspaces=false,
  showtabs=false,                  
  tabsize=2
}
\newcommand{\ie}{\textit{i.e.,}~}

\def\Snospace~{\S{}}

\definecolor{coco1}{HTML}{D9E4EC}
\definecolor{coco2}{HTML}{B7CFDC}
\definecolor{coco3}{HTML}{6AABD2}
\definecolor{coco4}{HTML}{385E72}
\hypersetup{
    colorlinks=true,
    linkcolor=coco3,
    filecolor=coco4,      
    urlcolor=coco3,
    citecolor=coco3,
}

\usepackage[frozencache,cachedir=minted-cache]{minted}

\title{Non-literal Understanding of Number Words by Language Models}
 
\author{
  {\large \bf Polina Tsvilodub$^\dagger$$^*$} \\
  Department of Linguistics, \\
  University of T\"ubingen, Germany
  \And {\large \bf Kanishk Gandhi$^\dagger$} \\
  Department of Computer Science, \\
  Stanford University
\And{\large \bf Haoran Zhao$^\dagger$} \\
  University of Washington
  \AND{\large \bf Jan-Philipp Fr\"anken} \\
  Department of Psychology, \\
  Stanford University
  \And {\large \bf Michael Franke} \\
  Department of Linguistics, \\
  University of T\"ubingen, Germany
  \And {\large \bf Noah D. Goodman} \\
  Departments of Psychology \& \\Computer Science, \\
  Stanford University
  }
 
\begin{document}
\maketitle
\begingroup
\renewcommand\thefootnote{}\footnotetext{$\dagger$ These authors contributed equally to this work.\\ \indent \indent $^*$polina.tsvilodub@uni-tuebingen.de}
\endgroup
\begin{abstract}
Humans naturally interpret numbers non-literally, effortlessly combining context, world knowledge, and speaker intent. We investigate whether large language models (LLMs)  interpret numbers similarly, focusing on hyperbole and pragmatic halo effects. Through systematic comparison with human data and computational models of pragmatic reasoning, we find that LLMs diverge from human interpretation in striking ways.
By decomposing pragmatic reasoning into testable components, grounded in the Rational Speech Act framework, we pinpoint where LLM processing diverges from human cognition --- not in prior knowledge, but in reasoning with it. 
This insight leads us to develop a targeted solution --- chain-of-thought prompting inspired by an RSA model makes LLMs' interpretations more human-like. Our work demonstrates how computational cognitive models can both diagnose AI-human differences and guide development of more human-like language understanding capabilities.

\end{abstract}

\section{Introduction}
\label{sec:intro}
\begin{table*}[t]

\centering
\resizebox{0.95\textwidth}{!}{
\begin{tabular}{p{3.2cm}p{18cm}}
\toprule
\textbf{Experiment} & \textbf{Prompt} \\
\midrule

\textbf{1: Hyperbole \& Halo} & 
In each scenario, two friends are talking about the price of an item.
Please read the scenarios carefully and provide the probability that the item has the described price.
Provide the estimates on a continuous scale between 0 and 1, where 0 stands for "impossible" and 1 stands for "extremely likely".
\textit{Daniel} bought a new \textit{electric kettle}. A friend asked him, ``Was it expensive?'' \textit{Daniel} said, ``It cost \textit{\$47}.''  Please provide the probability that the \textit{electric kettle} costs \textit{\$50}.
\\
\hline
\textbf{2: Affective Subtext} & In each scenario, a person has just bought an item and is talking to a friend about the price.
Please read the scenarios carefully and provide the probability that the person thinks that the item is expensive.
Provide the estimates on a continuous scale between 0 and 1, where 0 stands for "impossible" and 1 stands for "absolutely certain".
\textit{Daniel} bought a new \textit{electric kettle}. It cost \textit{\$47}. A friend asked him, ``Was it expensive?'' \textit{Daniel} said, ``It cost \textit{\$47}.''   Please provide the probability that \textit{Daniel} thinks that the \textit{electric kettle} is expensive. 
\\
\hline 
\textbf{3a: Price Prior} & Each scenario is about the price of an item.
Please read the scenarios carefully and provide the probability that someone buys the item with the given price.
Provide the estimates on a continuous scale between 0 and 1, where 0 stands for "impossible" and 1 stands for "extremely likely". 
\textit{Daniel} bought a new \textit{electric kettle}. It cost \textit{\$50}. Please provide the probability that someone buys the \textit{electric kettle} with this price.
\\
\hline 
\textbf{3b: Affect Prior} & In each scenario, someone has just bought an item.
Please read the scenarios carefully and provide the probability that the buyer thinks that the item is expensive.
Provide the estimates on a continuous scale between 0 and 1, where 0 stands for "impossible" and 1 stands for "absolutely certain".
\textit{Daniel} bought a new \textit{electric kettle}. It cost \textit{\$50}.  Please provide the probability that the buyer thinks that the \textit{electric kettle} is expensive.
\\
\bottomrule
\end{tabular}}
\caption{
Example prompts used in each experiment. The constant sentences were used as the system prompt. \textit{Italicized} segments varied in each trial.}
\label{tab:prompts}
\end{table*}

A friend exclaims, ``This coffee cost me a million dollars!'' We instantly understand the intended meaning: the coffee was surprisingly expensive (but not a million dollars). Humans often \emph{interpret words non-literally}, effortlessly integrating context, world knowledge, and speaker intent to grasp the meaning behind expressions \citep{gibbs2006figurative}. As large language models (LLMs) become increasingly integrated into our daily lives, three crucial questions emerge: 1) Do LLMs understand literal and non-literal utterances as humans do? 2) Can we use computational models of human cognition to systematically analyze how LLMs interpret non-literal utterances? 3) Can we use cognitive models of human pragmatic language understanding to guide LLMs to interpret meaning in a more human-like way?

In this work, we address these questions by focusing on two common phenomena in the interpretation of \textit{number} words: \textit{hyperbole}, the deliberate use of extreme numerical exaggeration to convey emotion or emphasis (see Ex.~\ref{ex:hyperbole}), and the \textit{pragmatic halo effect}, the tendency to interpret round numbers imprecisely (Ex.~\ref{ex:halo}) and sharp numbers precisely (Ex.~\ref{ex:exact}) \citep{lasersohn1999pragmatic,Krifka2007:Approximate-Int}: 

\ex.\label{ex1} Bob bought a kettle. Bob said: 
\a.\label{ex:hyperbole} `It cost \$10000.' $\leadsto$ \textit{Too expensive.} \hfill (hyperbole)
\b.\label{ex:halo} `It cost \$50.' $\leadsto$ \textit{It cost around \$50.} \hfill (imprecise)
\c.\label{ex:exact} `It cost \$48.' $\leadsto$ \textit{It cost exactly \$48.} \hfill (exact)

Language models trained to auto-regressively predict the next word and subsequently fine-tuned through human feedback have produced impressive performance in many areas \citep[among many others]{srivastava2023-BIGbench}. However, it remains unclear to what extent such training leads to nuanced distinction of literal and non-literal language in LLMs.
Recent work has explored non-literal language interpretation in LLMs, from metaphor comprehension \citep{ tong-etal-2021-recent,liu-etal-2022-testing,carenini2023large, prystawski2023psychologically} to pragmatic inference \citep{ jeretic-etal-2020-natural, jian2024llms, ruis2024goldilocks}. 
While benchmarking efforts have revealed persistent gaps between human and model performance \citep{sravanthi-etal-2024-pub}, we still lack a comprehensive understanding of when and why language models fail at interpreting non-literal language. Understanding these limitations is crucial both for improving models and for insights into how meaning is captured through large-scale language modeling.

To investigate whether LLMs interpret number words in a human-like way, as in Example~\ref{ex1}, we compare their  interpretations with human judgment data from \citet{kao2014nonliteral}. Specifically, we elicit LLMs' likelihood estimates for different prices given an uttered number, allowing us to compute the probability of hyperbolic interpretation (\ie interpreting the price as lower than the stated amount). 

The cognitive model of non-literal language interpretation in \citet{kao2014nonliteral}, suggests that human interpretation is driven by prior knowledge interacting with inferences about the goals of the speaker. We find an interesting disconnect --- while LLMs demonstrate human-like prior knowledge about typical prices and what constitutes ``expensive'', they tend toward more literal interpretations of numerical expressions. This suggests that despite having acquired accurate world knowledge through training, LLMs may lack the pragmatic reasoning mechanisms that humans use to bridge between literal meanings and intended interpretations. 

We then explore whether insights from cognitive science toward more human-like interpretations of hyperbole and pragmatic halo, comparing two approaches: cognitive model-inspired chain-of-thought prompting and direct implementation of computational reasoning steps with an LLM. Through both methods—providing explicit reasoning chains and implementing Rational Speech Act framework \citep{goodman2016pragmatic} computations—we demonstrate that LLMs can achieve more human-like interpretations of non-literal language.

\section{Pragmatic Number Interpretation in Humans}
\label{sec:related-work}

Our work builds on the computational cognitive model developed by \citet{kao2014nonliteral} in the Rational Speech Act (RSA) framework, which explains human interpretation of hyperbolic numerical expressions in terms of reasoning about the speaker's  communicative intent and prior world knowledge.
Specifically, the RSA framework models pragmatic communication as recursive rational reasoning between speakers and listeners \citep{goodman2016pragmatic, degen2023rational}. In the basic RSA model, a pragmatic speaker $S_1$ chooses utterances $u$ to inform a literal listener $L_0$ of a meaning $m$, minimizing the listener's surprisal: 
$$S_1(u \mid m) = \frac{\exp(\log(P(m \mid [\![u]\!])-C(u)))}{\sum_{u'} \exp(\log( P(m \mid [\![u']\!] ) - C(u')))} $$
where $C(u)$ is the cost of the utterance and $[\![u]\!]$ is the set of meanings compatible with $u$. A pragmatic listener $L_1$ then performs Bayesian inference over possible meanings by reasoning about this speaker:
$$ L_1(m \mid u)\propto S_1(u \mid m) P(m)$$
where $P(m)$ is the prior probability of a meaning.

To model hyperbolic interpretations like in our coffee example, \citet{kao2014nonliteral} extend this framework to capture how a single utterance can convey multiple meanings. Their extended model represents a multi-dimensional meaning space where an utterance about price conveys both the actual price state $s$ (e.g., the literal cost of the coffee) and the speaker's affect $a$ (e.g., that it was surprisingly expensive). The model also incorporates different communicative goals $g$, allowing the speaker to emphasize either or both of these dimensions: %\mf{best replace $L_0$ here as well if you followed my suggestion before for vanilla RSA}
$$S_1 (u \mid s, a, g) \propto \sum_{s', a'} \delta_{g(s, a) = g(s', a')}P(s', a' \mid [\![u]\!]) \cdot e^{-c(u)}$$
The pragmatic listener then interprets the utterance through joint inference over the speaker's goal and intended meaning:
$$L_1(s, a \mid u) \propto \sum_{g} S_1(u \mid s, a, g) P_{S}(s) P_{A}(a\mid s) P_{G}(g) $$
where $P_{S}$ represents prior beliefs about prices (e.g., how much coffee typically costs), $P_{A}$ captures the relationship between prices and affect (e.g., when a coffee price would be considered exasperating), and $P_{G}$ represents the prior over different communicative goals, assumed to be uniform. \citet{kao2014nonliteral} showed that this model successfully captures how humans interpret both hyperbolic expressions and the pragmatic differences between round and precise numbers, with model predictions strongly correlating with human judgments.

To explore whether the RSA model can guide LLMs toward more human-like interpretations, we develop a chain-of-thought (CoT) prompt that explicitly walks through key reasoning steps: considering possible speaker intentions, evaluating prior price expectations, and interpreting the utterance accordingly. We demonstrate this reasoning process with an example item (see supplementary) before eliciting the model's interpretation.

\section{Experiments}
\label{sec:experiments}

\renewcommand{\arraystretch}{1.5}
\begin{table*}[ht!]
    \centering
    \small
    \begin{tabularx}{\textwidth}{l*{7}{>{\centering\arraybackslash}X}}
        \toprule    
        \multicolumn{1}{l}{ } & \multicolumn{2}{c}{GPT-4o-mini} & \multicolumn{2}{c}{Claude-3.5 Sonnet} & \multicolumn{2}{c}{Gemini-1.5-pro} \\
        
        \cmidrule(lr){2-3} \cmidrule(lr){4-5} \cmidrule(lr){6-7}
        
        Prompt & 0-shot & 1-shot RSA CoT & 0-shot & 1-shot RSA CoT & 0-shot & 1-shot RSA CoT\\
        \midrule
         $R$ with humans & 0.41 & 0.579 & 0.528 & 0.558 & 0.365 & 0.603 \\
        \bottomrule
    \end{tabularx}
    \caption{Correlations between human data and LLM predictions of probabilities of all utterance-meaning ($(s, u)$) pairs. 0-shot indicates correlations of human results with LLM results under 0-shot prompting, 1-shot RSA CoT indicates correlations of human results with LLM results under one-shot RSA-based CoT prompting.
    \label{tab:cor_table}}
\end{table*}
 
We closely follow the procedure and the scenarios presented in \cite{kao2014nonliteral}, about three daily items: an electric kettle, a watch, and a laptop. We study three LLMs in our experiments: GPT-4o-mini, Claude-3.5-sonnet, and and Gemini-1.5-pro. We sample responses from the LLMs with temperature $\tau=1$ for $n=10$ times for each query and average predictions across runs.\footnote{All materials and data are available at \href{https://sites.google.com/view/pragmatic-lms/home}{https://sites.google.com/view/pragmatic-lms/home}.}

\begin{figure*}[t]
    \centering
    \includegraphics[width=0.8\textwidth]{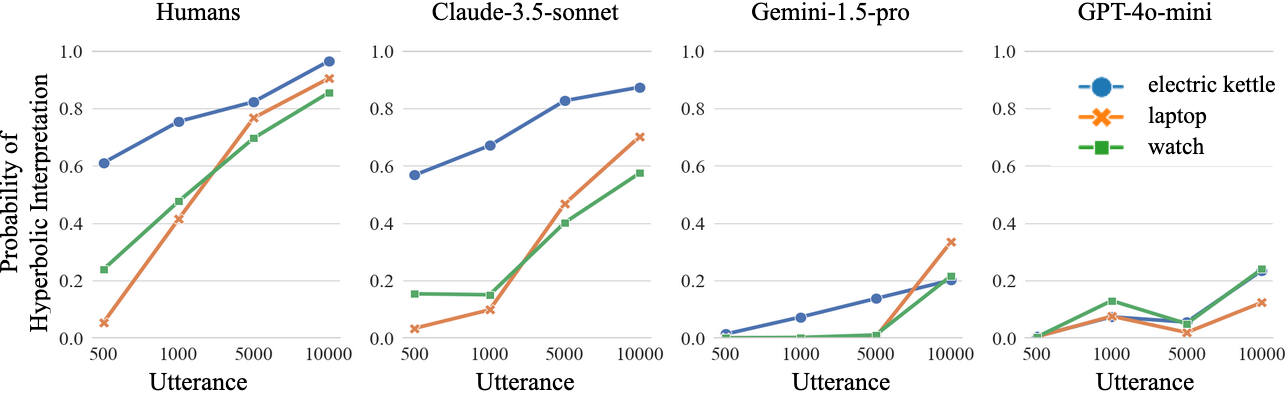} 
    \caption{Probability of hyperbolic interpretation, i.e., $u > s$, averaged over sharp and round values of $u$. 
    }
    \label{fig:hyperbole}
\end{figure*}

\begin{figure*}[tbp]
    \centering
    \includegraphics[width=0.9\textwidth]{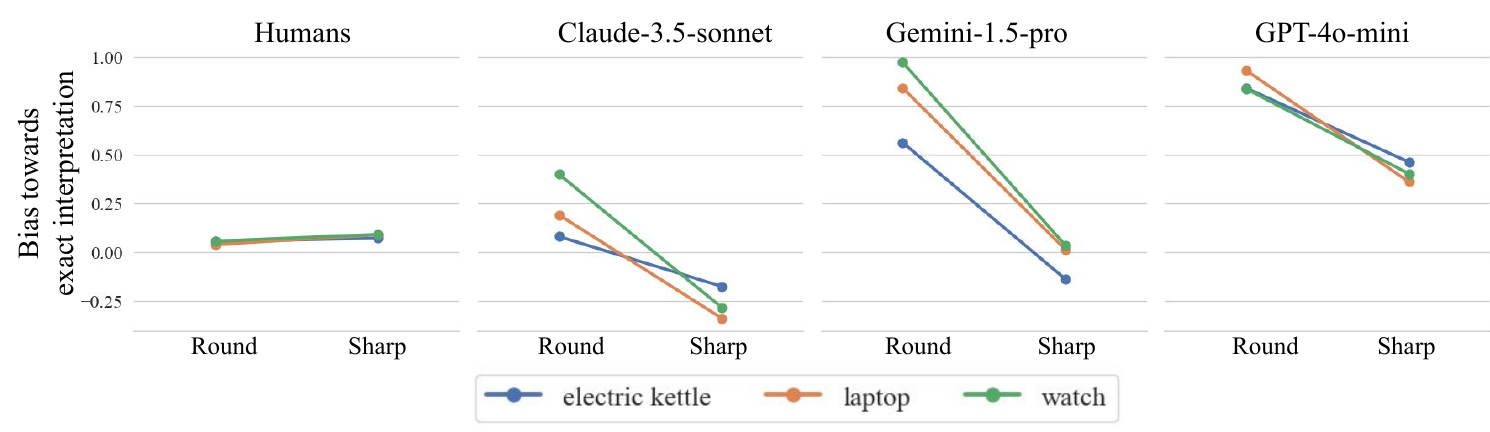} 
    \caption{Bias towards pragmatic halo interpretation, calculated by subtracting the probability of a fuzzy interpretation from the probability of the exact interpretation.
    } 
    \label{fig:halo}
\end{figure*}

\subsection{Experiment 1: Hyperbole and Halo}
In this experiment, we examine how LLMs interpret price-related utterances, comparing their behavior to human patterns. For hyperbole understanding, we expect LLMs to assign lower probabilities to literal interpretations when they are contextually implausible—for instance, the likelihood of a literal \$10,000 interpretation should be low when discussing an electric kettle's price. To assess pragmatic halo effects, we compare interpretations of sharp versus round numbers, hypothesizing that exact interpretations should be more probable for precise utterances (e.g., ``\$51'') than for round ones (e.g., ``\$50''). To quantify human-likeness, we correlate the LLMs' probability distributions over different price states $s$ given an utterance $u$ with human judgments collected by \citet{kao2014nonliteral}, for both hyperbolic and halo effects.
%Further, the likelihood of hyperbole should increase with higher uttered prices. 

\paragraph{Materials and Procedure.}  
The prompts for all experiments were kept as close as possible to original human experiments. Following \citet{kao2014nonliteral}, we used the following sets of price states and utterances $U = S$ = \{50 + k, 500 + k, 1000 + k, 5000 + k, 10000 + k\}, with $k \in \{0, 1, 2, 3\}$ to create exact and round prices.
The same procedure was applied to all three items.
For each utterance with $u \in U$ of the form ``It cost \$\textit{u}.'', the LLMs were prompted to predict the probability that the item had each price $s \in S$. The probabilities were then renormalized over $S$ for each $u \in U$. 

First, we use a \textit{zero-shot} prompt to generate probabilities of possible price states, given the utterance; examples are presented in \autoref{tab:prompts} (1: Hyperbole \& Halo). 
Second, we guide the models using a \textit{one-shot chain-of-thought (CoT) prompt} \citep{nye2021show,wei2022chain}. We construct this prompt by translating the computational steps of the RSA model into natural language reasoning for an example scenario \citep[following][]{prystawski2023psychologically}. This RSA-inspired chain-of-thought is appended to our original system prompt as shown in \autoref{tab:prompts} (1), followed by the context and target task.\footnote{The full CoT prompt is in our supplementary materials.} 

\paragraph{Results.}
Results from our zero-shot evaluation reveal significant disparities between LLM and human interpretations of price-related utterances, as shown in \autoref{tab:cor_table}. 
When examining correlations with human data, we find that LLMs generally default to literal interpretations, with even the best-performing model (Claude-3.5-sonnet) achieving only moderate correlation. Different models exhibited distinct behavioral patterns: GPT-4o-mini tended to assign inflated probabilities to individual utterance-meaning pairs, while Gemini-1.5-pro generally exhibited a bimodal distribution of ratings at the ends of the scale.\footnote{0-shot distributional results are in the supplementary materials.}
For hyperbolic interpretation (\autoref{fig:hyperbole}), we analyzed the probability of hyperbolic meaning by summing probabilities of states where the utterance exceeds the true state ($u > s$), averaging across both round and sharp values (e.g., \$50 and \$49). Only Claude-3.5-sonnet demonstrated a consistent pattern of increasing hyperbolic interpretation probability with higher utterances, though this pattern matched human behavior most closely in the electric kettle domain (cf. \autoref{fig:hyperbole}, left). Other models consistently underestimated hyperbolic interpretations compared to human benchmarks.

The halo effect analysis (\autoref{fig:halo}) revealed an even more striking divergence from human behavior. We quantified halo bias by calculating the difference between exact interpretation probabilities ($s = u$) and fuzzy interpretation probabilities ($s \neq u$ and $s \in [u-3, u+3]$).
While humans showed a small preference for exact interpretations with sharp numbers, LLMs displayed a large effect in the opposite direction, favoring exact interpretations for round numbers. 
These findings demonstrate that contemporary LLMs fail to capture human-like pragmatic reasoning in hyperbole and halo interpretation.
\paragraph{RSA-like Chain-of-Thought.} We explored if a one-shot Chain of Thought (CoT) prompt, describing the computational process of an RSA model would make LLM responses more human-like. We found that the 1-shot RSA prompt improved correlations between model predictions and human data for GPT-4o-mini and Gemini-1.5-pro (\autoref{tab:cor_table}, 1-shot RSA CoT). 
Interestingly, additional ablation studies showed that prompts explaining only a few components of the RSA model (e.g., mentioning only speaker goals or only price priors) achieved comparable  improvements (see supplementary material for details).
These ablation results suggest that while RSA-inspired prompting can improve LLM performance, the minimal components sufficient for this improvement differ from the full computational process required to explain human behavior.

\subsection{Experiment 2: Affective Subtext}
\begin{figure*}[tbp]
    \centering
    \includegraphics[width=0.8\textwidth]{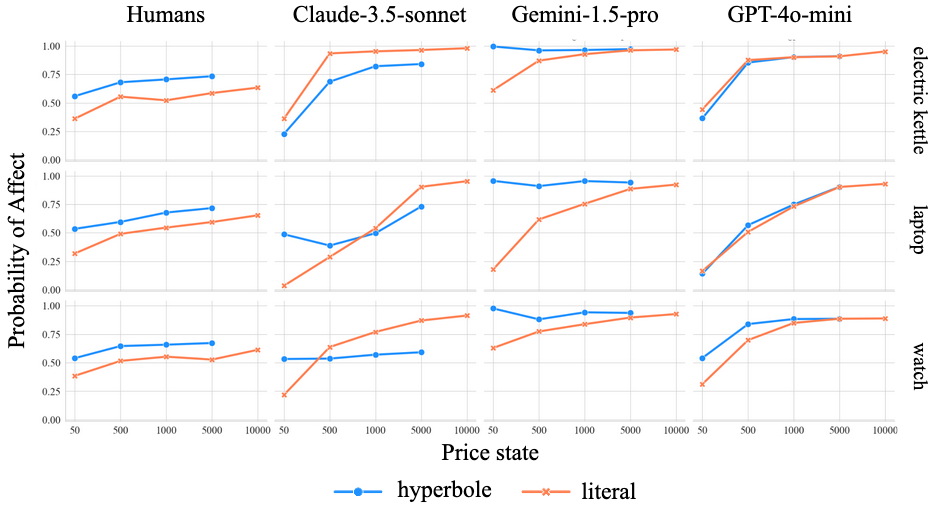} 
    \caption{Probability of speaker affect (y-axis), given a price $s$ and an utterance $u$, predicted by LLMs with zero-shot prompting (with $\tau=1$) and by humans in Experiment 2 (columns), for different items (rows). Affect is rated by for literal utterances where $u=s$ and hyperbolic utterances where $u > s$. 
    }
    \label{fig:expt2}
\end{figure*}

Next, we assess the inferred probability that a speaker thinks an item is strikingly expensive (\ie expressed affect), given the description of the true price $s$ and the speaker's statement $u$. If the LLMs interpret hyperbole as conveying affect, the likelihood of affect will be higher for hyperbolic utterances (i.e., $u > s$) than for literal utterances (i.e., $u = s$).
\paragraph{Materials and Procedure.} 
We use the same procedure as in Experiment 1 to retrieve the probability of affect, given a \textit{zero-shot prompt} as exemplified in \autoref{tab:prompts} (2). We use the same sets $S$ and $U$ as in Experiment 1.
Following the original experiment, we then round all the states, since we do not predict differences in affect between round and sharp utterances, and calculate the average probabilities of affect for literal utterances (where $s=u$) and hyperbolic utterances (where $u > s$). %We don't use trials where $u < s$.
\paragraph{Results.}
Results are shown in \autoref{fig:expt2}. While humans in the experiment from \citet{kao2014nonliteral} robustly inferred a distinction between literal and hyperbolic utterances, predicting higher probability of affect given hyperbolic than literal utterances (leftmost facets), LLMs did not. GPT-4o-mini overestimated affect compared to humans, mostly collapsing across literal and hyperbolic utterances. Gemini-1.5-pro treated literal and non-literal utterances more distinctly, but also overestimated affect. Claude was more conservative than humans for hyperbolic utterances, but overestimated affect for literal utterances, with an opposite pattern to humans.

Overall, LLMs did not capture human patterns well when predicting affect probability, given the true price, $s$, and the uttered price, $u$. This suggests that LLMs do not map between utterances and affect in a human-like way. This failure may be ancillary to the failure to capture hyperbolic interpretations or may reflect further difficulties with affect (though see \citet{gandhi2024human}, where LLMs demonstrate some human-like affective cognition in other contexts).

\begin{table}[ht!]
\centering
\begin{tabular}{llll}
\toprule
 \textbf{LLM} & GPT & Claude & Gemini  \\ \midrule
 Price prior  & 0.889 & 0.93 &  0.92  \\ 
 Affect prior & 0.95 & 0.973 & 0.779 \\
 \bottomrule
\end{tabular}
\caption{Correlations of human judgments and different LLM predictions, for price and conditional affect prior probabilities of different prices, across items.
These priors were used to fit the respective LM-RSA models in Experiment 3. \label{tab:priors}}
\end{table}

\subsection{Experiment 3: Price and Affect Priors}

Given that Experiments 1 and 2 revealed significant differences between LLM and human behavior in processing hyperbole, halo effects, and affective predictions, we designed Experiment 3 to investigate a potential root cause: the accuracy of LLMs' prior knowledge about price distributions and price-affect relationships. Previous work by \citet{kao2014nonliteral} has demonstrated that these priors strongly influence human pragmatic inference. Our experiment focuses on two key aspects: (1) the probability distributions that LLMs assign to different price states for each item, and (2) their predictions of affective responses conditional on item prices. By comparing these model-generated priors against human benchmarks, we can assess whether deficiencies in base knowledge might explain the models' poor performance in pragmatic reasoning tasks. 

Strong correlations between LLM and human probability estimates would suggest that pragmatic failures stem from reasoning mechanisms rather than knowledge gaps, while weak correlations would point to fundamental limitations in the models' basic priors about price and affect. By fitting an RSA model using LLM-generated priors, we can quantitatively assess the models' internal consistency—specifically, whether their predictions align with their own stated priors. This analysis provides a systematic approach for isolating reasoning deficiencies independent of the accuracy of the priors themselves. To enable even more precise analysis of the LLMs' priors and reasoning, we additionally fit an RSA model incorporating both LLM-generated priors and conditional utterance likelihoods. If both RSA models demonstrate strong alignment with human data, the reasoning deficiency could be localized specifically to the model's ability to reason about a speaker's intentions.

\paragraph{Materials and Procedure.} 
We use a fixed set of price states $S$ = \{50 + k, 500 + k, 1000 + k, 5000 + k, 10000 + k\}, where $k$ was  selected from the set \{0, 1\}. 
For price priors, we retrieve LLM predictions with the \textit{zero-shot} prompt asking the LLM to assess the probability of each price $s \in S$ (\autoref{tab:prompts} (3a)). We renormalize the predictions over all prices $S$ for each item. 
To retrieve priors over affect, we prompt the LLM \textit{zero-shot} to provide the probability that a person thinks an item is expensive, given the price $s\in S$ of that item (\autoref{tab:prompts} (3b)). We treat the predictions for each price $s$ as the probability of affect $P(a \mid s)$. 
To retrieve the conditional probabilities of different utterances, we construct a prompt verbalizing the state and different goals of the RSA speaker $S_1$.\footnote{Details can be found in the supplementary materials.}  
Finally, to investigate to which extent LLMs are consistent with their own priors, we use the predicted \textit{LLM priors} for parameterizing the $P_{S}$ and $P_A$ in the RSA model.
We call the resulting models \textit{LM-RSA} and implement them using a probabilistic programming language WebPPL \citep{dippl}. 

\begin{figure*}[tbp]
    \centering
    \includegraphics[width = 0.75\textwidth]{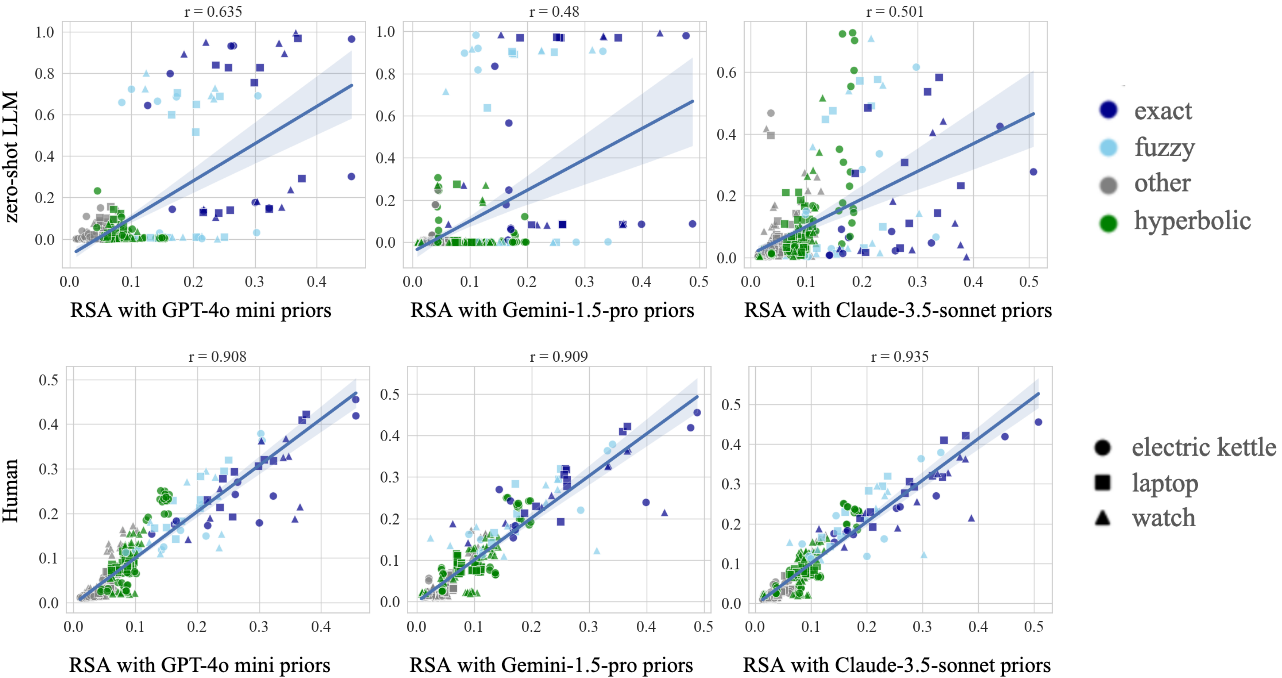} 
    \caption{Correlation of predicted probabilities of each pair of $(u, s)$. The plots in the upper panel show predictions of the RSA model with LLM priors (x-axis) against predictions of the same LLM under zero-shot prompting (y-axis). The plot in the lower panel shows predictions of the RSA model with LLM priors (x-axis) against human results (y-axis).}
    \label{fig:lm-rsa-claude-gemini}
\end{figure*}

\paragraph{Results.}
We find that LLM-predicted priors show strong correlation with human data ($r>0.7$) across price distributions and affect relationships (\autoref{tab:priors}). This indicates that LLMs possess the prior knowledge that should, in principle, enable them to perform human-like pragmatic inference for both hyperbole and halo effects. We observed a systematic relationship between prior accuracy and zero-shot performance: models with stronger correlations to human priors (progressing from Gemini to Claude) demonstrated correspondingly better zero-shot performance. However, this alignment in prior knowledge, while necessary, proved insufficient to guarantee human-like pragmatic reasoning under prompt-based evaluation.

 Do deviations in interpretation nonetheless derive from the small deviations in prior knowledge?
 We compared models' zero-shot behavior against predictions from RSA models fitted to each LLM's own priors (\autoref{fig:lm-rsa-claude-gemini}, upper panel). The results exposed a fundamental inconsistency: LLMs' zero-shot predictions showed relatively weak correlations with their corresponding RSA predictions, with even the best-performing model (GPT-4o-mini) achieving only moderate correlation (0.635). Yet, these same RSA models showed strong correlations with human judgments (\autoref{fig:lm-rsa-claude-gemini}, lower panel) showing that LLM priors are aligned with human priors for price and affect  ($R>0.9$). The LM-RSA models which also included LLM-generated utterance probabilities showed comparable correlations with human judgments  ($R>0.7$).\footnote{Full results are provided in the supplement.}
 This suggests that the challenge in achieving human-like pragmatic reasoning lies not in the models' priors, but in their ability to systematically apply them during inference.

\section{Discussion}
\label{sec:discussion}
 We compared human interpretation data for numerical utterances to LLMs' interpretations, finding substantial differences.
This manifests in LLMs' tendency toward literal interpretations, reversed halo effects (preferring exact interpretations for round rather than sharp numbers; Exp.~1), and inconsistent affect attribution between literal and hyperbolic utterances (Exp.~2), despite human-like prior representations (Exp.~3).
These findings point to a disconnect in LLM pragmatic reasoning --- despite possessing accurate prior knowledge about prices, affect and utterance probabilities --- and despite this knowledge being structured in a way that could support human-like inference when processed through an RSA framework --- LLMs fail to consistently leverage this information when directly prompted to make pragmatic interpretations. 

Our findings highlight an important methodological contribution for understanding LLM behaviors: by systematically decomposing pragmatic reasoning into testable components (priors, affect mappings, utterance likelihoods, and interpretations), we can precisely locate differences between human and AI reasoning. 
This approach extends beyond traditional behavioral comparisons, allowing us to identify whether differences stem from knowledge gaps or reasoning mechanisms. Such detailed cognitive modeling approaches may prove valuable for understanding other aspects of LLM behavior, particularly in cases where surface-level performance masks deeper processing differences from human cognition.
Importantly, our results demonstrate that cognitively-inspired chain-of-thought prompting can help bridge this gap between knowledge and application. We achieved improved correlations with human judgments by decomposing the RSA model's computational steps into natural language reasoning chains. This success suggests that while LLMs may not naturally develop human-like pragmatic reasoning through training alone, they can successfully implement such reasoning when given appropriate computational frameworks that mirror human cognitive processes.

Future research could address important follow-up questions. For instance, frequency effects of different non-literal expressions \citep[cf.][]{mccoy2024embers} or potential training modifications to help LLMs better integrate their prior knowledge and context when interpreting hyperbole could be analyzed. Identifying factors that influence how LLMs apply this knowledge in context is also an open question. Our supplementary materials report exploratory analyses that begin to probe these questions through variations in prompting of the models.

Ultimately, our work demonstrates that evaluating LLMs through the lens of cognitive modeling provides a nuanced understanding of how these models deviate from human understanding. By integrating LLMs with cognitive models of pragmatic language use, we can both critically assess the models' internal consistency and provide a framework for improving their performance in interpreting non-literal language.

\section{Acknowledgments}
PT and MF acknowledge support by the state of Baden-W\"urttemberg through bwHPC and the German Research Foundation (DFG) through grant INST 35/1597-1 FUGG. 
MF is a member of the Machine Learning Cluster of Excellence at University of T\"ubingen, EXC number 2064/1 – Project number 39072764. KG was supported by an HAI-SAP Grant and NSF Expeditions Grant Award Number (FAIN) 1918771.
\bibliographystyle{apacite}

\setlength{\bibleftmargin}{.125in}
\setlength{\bibindent}{-\bibleftmargin}

\bibliography{CogSci_Template}

\begin{thebibliography}{}

\bibitem [\protect \citeauthoryear {%
Carenini%
, Bodot%
, Bischetti%
, Schaeken%
\BCBL {}\ \BBA {} Bambini%
}{%
Carenini%
\ \protect \BOthers {.}}{%
{\protect \APACyear {2023}}%
}]{%
carenini2023large}
\APACinsertmetastar {%
carenini2023large}%
\begin{APACrefauthors}%
Carenini, G.%
, Bodot, L.%
, Bischetti, L.%
, Schaeken, W.%
\BCBL {}\ \BBA {} Bambini, V.%
\end{APACrefauthors}%
\unskip\
\newblock
\APACrefYearMonthDay{2023}{}{}.
\newblock
{\BBOQ}\APACrefatitle {Large language models behave (almost) as rational speech actors: Insights from metaphor understanding} {Large language models behave (almost) as rational speech actors: Insights from metaphor understanding}.{\BBCQ}
\newblock
\BIn{} \APACrefbtitle {NeurIPS 2023 workshop: Information-Theoretic Principles in Cognitive Systems.} {Neurips 2023 workshop: Information-theoretic principles in cognitive systems.}
\PrintBackRefs{\CurrentBib}

\bibitem [\protect \citeauthoryear {%
Degen%
}{%
Degen%
}{%
{\protect \APACyear {2023}}%
}]{%
degen2023rational}
\APACinsertmetastar {%
degen2023rational}%
\begin{APACrefauthors}%
Degen, J.%
\end{APACrefauthors}%
\unskip\
\newblock
\APACrefYearMonthDay{2023}{}{}.
\newblock
{\BBOQ}\APACrefatitle {The rational speech act framework} {The rational speech act framework}.{\BBCQ}
\newblock
\APACjournalVolNumPages{Annual Review of Linguistics}{9}{1}{519--540}.
\PrintBackRefs{\CurrentBib}

\bibitem [\protect \citeauthoryear {%
Gandhi%
\ \protect \BOthers {.}}{%
Gandhi%
\ \protect \BOthers {.}}{%
{\protect \APACyear {2024}}%
}]{%
gandhi2024human}
\APACinsertmetastar {%
gandhi2024human}%
\begin{APACrefauthors}%
Gandhi, K.%
, Lynch, Z.%
, Fr{\"a}nken, J\BHBI P.%
, Patterson, K.%
, Wambu, S.%
, Gerstenberg, T.%
\BDBL {}Goodman, N\BPBI D.%
\end{APACrefauthors}%
\unskip\
\newblock
\APACrefYearMonthDay{2024}{}{}.
\newblock
{\BBOQ}\APACrefatitle {Human-like affective cognition in foundation models} {Human-like affective cognition in foundation models}.{\BBCQ}
\newblock
\APACjournalVolNumPages{arXiv preprint arXiv:2409.11733}{}{}{}.
\PrintBackRefs{\CurrentBib}

\bibitem [\protect \citeauthoryear {%
Gibbs~Jr%
\ \BBA {} Colston%
}{%
Gibbs~Jr%
\ \BBA {} Colston%
}{%
{\protect \APACyear {2006}}%
}]{%
gibbs2006figurative}
\APACinsertmetastar {%
gibbs2006figurative}%
\begin{APACrefauthors}%
Gibbs~Jr, R\BPBI W.%
\BCBT {}\ \BBA {} Colston, H\BPBI L.%
\end{APACrefauthors}%
\unskip\
\newblock
\APACrefYearMonthDay{2006}{}{}.
\newblock
{\BBOQ}\APACrefatitle {Figurative language} {Figurative language}.{\BBCQ}
\newblock
\BIn{} \APACrefbtitle {Handbook of psycholinguistics} {Handbook of psycholinguistics}\ (\BPGS\ 835--861).
\newblock
\APACaddressPublisher{}{Elsevier}.
\PrintBackRefs{\CurrentBib}

\bibitem [\protect \citeauthoryear {%
Goodman%
\ \BBA {} Frank%
}{%
Goodman%
\ \BBA {} Frank%
}{%
{\protect \APACyear {2016}}%
}]{%
goodman2016pragmatic}
\APACinsertmetastar {%
goodman2016pragmatic}%
\begin{APACrefauthors}%
Goodman, N\BPBI D.%
\BCBT {}\ \BBA {} Frank, M\BPBI C.%
\end{APACrefauthors}%
\unskip\
\newblock
\APACrefYearMonthDay{2016}{}{}.
\newblock
{\BBOQ}\APACrefatitle {Pragmatic language interpretation as probabilistic inference} {Pragmatic language interpretation as probabilistic inference}.{\BBCQ}
\newblock
\APACjournalVolNumPages{Trends in cognitive sciences}{20}{11}{818--829}.
\PrintBackRefs{\CurrentBib}

\bibitem [\protect \citeauthoryear {%
Goodman%
\ \BBA {} Stuhlm\"{u}ller%
}{%
Goodman%
\ \BBA {} Stuhlm\"{u}ller%
}{%
{\protect \APACyear {2014}}%
}]{%
dippl}
\APACinsertmetastar {%
dippl}%
\begin{APACrefauthors}%
Goodman, N\BPBI D.%
\BCBT {}\ \BBA {} Stuhlm\"{u}ller, A.%
\end{APACrefauthors}%
\unskip\
\newblock
\APACrefYearMonthDay{2014}{}{}.
\newblock
\APACrefbtitle {{The Design and Implementation of Probabilistic Programming Languages}.} {{The Design and Implementation of Probabilistic Programming Languages}.}
\newblock
\APAChowpublished {\url{http://dippl.org}}.
\newblock
\APACrefnote{Accessed: 2025-1-25}
\PrintBackRefs{\CurrentBib}

\bibitem [\protect \citeauthoryear {%
Jeretic%
, Warstadt%
, Bhooshan%
\BCBL {}\ \BBA {} Williams%
}{%
Jeretic%
\ \protect \BOthers {.}}{%
{\protect \APACyear {2020}}%
}]{%
jeretic-etal-2020-natural}
\APACinsertmetastar {%
jeretic-etal-2020-natural}%
\begin{APACrefauthors}%
Jeretic, P.%
, Warstadt, A.%
, Bhooshan, S.%
\BCBL {}\ \BBA {} Williams, A.%
\end{APACrefauthors}%
\unskip\
\newblock
\APACrefYearMonthDay{2020}{{\APACmonth{07}}}{}.
\newblock
{\BBOQ}\APACrefatitle {Are Natural Language Inference Models {IMPPRESsive}? {L}earning {IMPlicature} and {PRESupposition}} {Are natural language inference models {IMPPRESsive}? {L}earning {IMPlicature} and {PRESupposition}}.{\BBCQ}
\newblock
\BIn{} D.~Jurafsky, J.~Chai, N.~Schluter\BCBL {}\ \BBA {} J.~Tetreault\ (\BEDS), \APACrefbtitle {Proceedings of the 58th Annual Meeting of the Association for Computational Linguistics} {Proceedings of the 58th annual meeting of the association for computational linguistics}\ (\BPGS\ 8690--8705).
\newblock
\APACaddressPublisher{Online}{Association for Computational Linguistics}.
\newblock
\begin{APACrefURL} \url{https://aclanthology.org/2020.acl-main.768} \end{APACrefURL}
\newblock
\begin{APACrefDOI} \doi{10.18653/v1/2020.acl-main.768} \end{APACrefDOI}
\PrintBackRefs{\CurrentBib}

\bibitem [\protect \citeauthoryear {%
Jian%
\ \BBA {} Siddharth%
}{%
Jian%
\ \BBA {} Siddharth%
}{%
{\protect \APACyear {2024}}%
}]{%
jian2024llms}
\APACinsertmetastar {%
jian2024llms}%
\begin{APACrefauthors}%
Jian, M.%
\BCBT {}\ \BBA {} Siddharth, N.%
\end{APACrefauthors}%
\unskip\
\newblock
\APACrefYearMonthDay{2024}{}{}.
\newblock
{\BBOQ}\APACrefatitle {Are LLMs good pragmatic speakers?} {Are llms good pragmatic speakers?}{\BBCQ}
\newblock
\APACjournalVolNumPages{arXiv preprint arXiv:2411.01562}{}{}{}.
\PrintBackRefs{\CurrentBib}

\bibitem [\protect \citeauthoryear {%
Kao%
, Wu%
, Bergen%
\BCBL {}\ \BBA {} Goodman%
}{%
Kao%
\ \protect \BOthers {.}}{%
{\protect \APACyear {2014}}%
}]{%
kao2014nonliteral}
\APACinsertmetastar {%
kao2014nonliteral}%
\begin{APACrefauthors}%
Kao, J\BPBI T.%
, Wu, J\BPBI Y.%
, Bergen, L.%
\BCBL {}\ \BBA {} Goodman, N\BPBI D.%
\end{APACrefauthors}%
\unskip\
\newblock
\APACrefYearMonthDay{2014}{}{}.
\newblock
{\BBOQ}\APACrefatitle {Nonliteral understanding of number words} {Nonliteral understanding of number words}.{\BBCQ}
\newblock
\APACjournalVolNumPages{Proceedings of the National Academy of Sciences}{111}{33}{12002-12007}.
\newblock
\begin{APACrefURL} \url{https://www.pnas.org/doi/abs/10.1073/pnas.1407479111} \end{APACrefURL}
\newblock
\begin{APACrefDOI} \doi{10.1073/pnas.1407479111} \end{APACrefDOI}
\PrintBackRefs{\CurrentBib}

\bibitem [\protect \citeauthoryear {%
Krifka%
}{%
Krifka%
}{%
{\protect \APACyear {2007}}%
}]{%
Krifka2007:Approximate-Int}
\APACinsertmetastar {%
Krifka2007:Approximate-Int}%
\begin{APACrefauthors}%
Krifka, M.%
\end{APACrefauthors}%
\unskip\
\newblock
\APACrefYearMonthDay{2007}{}{}.
\newblock
{\BBOQ}\APACrefatitle {Approximate Interpretation of Number Words: {A} Case for Strategic Communication} {Approximate interpretation of number words: {A} case for strategic communication}.{\BBCQ}
\newblock
\BIn{} G.~Bouma, I.~Kr\"{a}mer\BCBL {}\ \BBA {} J.~Zwarts\ (\BEDS), \APACrefbtitle {Cognitive Foundations of Interpretation} {Cognitive foundations of interpretation}\ (\BPGS\ 111--126).
\newblock
\APACaddressPublisher{Amsterdam}{KNAW}.
\PrintBackRefs{\CurrentBib}

\bibitem [\protect \citeauthoryear {%
Lasersohn%
}{%
Lasersohn%
}{%
{\protect \APACyear {1999}}%
}]{%
lasersohn1999pragmatic}
\APACinsertmetastar {%
lasersohn1999pragmatic}%
\begin{APACrefauthors}%
Lasersohn, P.%
\end{APACrefauthors}%
\unskip\
\newblock
\APACrefYearMonthDay{1999}{}{}.
\newblock
{\BBOQ}\APACrefatitle {Pragmatic halos} {Pragmatic halos}.{\BBCQ}
\newblock
\APACjournalVolNumPages{Language}{}{}{522--551}.
\PrintBackRefs{\CurrentBib}

\bibitem [\protect \citeauthoryear {%
Liu%
, Cui%
, Zheng%
\BCBL {}\ \BBA {} Neubig%
}{%
Liu%
\ \protect \BOthers {.}}{%
{\protect \APACyear {2022}}%
}]{%
liu-etal-2022-testing}
\APACinsertmetastar {%
liu-etal-2022-testing}%
\begin{APACrefauthors}%
Liu, E.%
, Cui, C.%
, Zheng, K.%
\BCBL {}\ \BBA {} Neubig, G.%
\end{APACrefauthors}%
\unskip\
\newblock
\APACrefYearMonthDay{2022}{{\APACmonth{07}}}{}.
\newblock
{\BBOQ}\APACrefatitle {Testing the Ability of Language Models to Interpret Figurative Language} {Testing the ability of language models to interpret figurative language}.{\BBCQ}
\newblock
\BIn{} M.~Carpuat, M\BHBI C.~de Marneffe\BCBL {}\ \BBA {} I\BPBI V.~Meza~Ruiz\ (\BEDS), \APACrefbtitle {Proceedings of the 2022 Conference of the North American Chapter of the Association for Computational Linguistics: Human Language Technologies} {Proceedings of the 2022 conference of the north american chapter of the association for computational linguistics: Human language technologies}\ (\BPGS\ 4437--4452).
\newblock
\APACaddressPublisher{Seattle, United States}{Association for Computational Linguistics}.
\newblock
\begin{APACrefURL} \url{https://aclanthology.org/2022.naacl-main.330/} \end{APACrefURL}
\newblock
\begin{APACrefDOI} \doi{10.18653/v1/2022.naacl-main.330} \end{APACrefDOI}
\PrintBackRefs{\CurrentBib}

\bibitem [\protect \citeauthoryear {%
McCoy%
, Yao%
, Friedman%
, Hardy%
\BCBL {}\ \BBA {} Griffiths%
}{%
McCoy%
\ \protect \BOthers {.}}{%
{\protect \APACyear {2024}}%
}]{%
mccoy2024embers}
\APACinsertmetastar {%
mccoy2024embers}%
\begin{APACrefauthors}%
McCoy, R\BPBI T.%
, Yao, S.%
, Friedman, D.%
, Hardy, M\BPBI D.%
\BCBL {}\ \BBA {} Griffiths, T\BPBI L.%
\end{APACrefauthors}%
\unskip\
\newblock
\APACrefYearMonthDay{2024}{}{}.
\newblock
{\BBOQ}\APACrefatitle {Embers of autoregression show how large language models are shaped by the problem they are trained to solve} {Embers of autoregression show how large language models are shaped by the problem they are trained to solve}.{\BBCQ}
\newblock
\APACjournalVolNumPages{Proceedings of the National Academy of Sciences}{121}{41}{e2322420121}.
\PrintBackRefs{\CurrentBib}

\bibitem [\protect \citeauthoryear {%
Nye%
\ \protect \BOthers {.}}{%
Nye%
\ \protect \BOthers {.}}{%
{\protect \APACyear {2021}}%
}]{%
nye2021show}
\APACinsertmetastar {%
nye2021show}%
\begin{APACrefauthors}%
Nye, M.%
, Andreassen, A\BPBI J.%
, Gur-Ari, G.%
, Michalewski, H.%
, Austin, J.%
, Bieber, D.%
\BDBL {}others%
\end{APACrefauthors}%
\unskip\
\newblock
\APACrefYearMonthDay{2021}{}{}.
\newblock
{\BBOQ}\APACrefatitle {Show your work: Scratchpads for intermediate computation with language models} {Show your work: Scratchpads for intermediate computation with language models}.{\BBCQ}
\newblock
\APACjournalVolNumPages{arXiv preprint arXiv:2112.00114}{}{}{}.
\PrintBackRefs{\CurrentBib}

\bibitem [\protect \citeauthoryear {%
Prystawski%
, Thibodeau%
, Potts%
\BCBL {}\ \BBA {} Goodman%
}{%
Prystawski%
\ \protect \BOthers {.}}{%
{\protect \APACyear {2023}}%
}]{%
prystawski2023psychologically}
\APACinsertmetastar {%
prystawski2023psychologically}%
\begin{APACrefauthors}%
Prystawski, B.%
, Thibodeau, P.%
, Potts, C.%
\BCBL {}\ \BBA {} Goodman, N.%
\end{APACrefauthors}%
\unskip\
\newblock
\APACrefYearMonthDay{2023}{}{}.
\newblock
{\BBOQ}\APACrefatitle {Psychologically-informed chain-of-thought prompts for metaphor understanding in large language models.} {Psychologically-informed chain-of-thought prompts for metaphor understanding in large language models.}{\BBCQ}
\newblock
\BIn{} \APACrefbtitle {Proceedings of the Annual Meeting of the Cognitive Science Society} {Proceedings of the annual meeting of the cognitive science society}\ (\BVOL~45).
\PrintBackRefs{\CurrentBib}

\bibitem [\protect \citeauthoryear {%
Ruis%
\ \protect \BOthers {.}}{%
Ruis%
\ \protect \BOthers {.}}{%
{\protect \APACyear {2024}}%
}]{%
ruis2024goldilocks}
\APACinsertmetastar {%
ruis2024goldilocks}%
\begin{APACrefauthors}%
Ruis, L.%
, Khan, A.%
, Biderman, S.%
, Hooker, S.%
, Rockt{\"a}schel, T.%
\BCBL {}\ \BBA {} Grefenstette, E.%
\end{APACrefauthors}%
\unskip\
\newblock
\APACrefYearMonthDay{2024}{}{}.
\newblock
{\BBOQ}\APACrefatitle {The goldilocks of pragmatic understanding: Fine-tuning strategy matters for implicature resolution by llms} {The goldilocks of pragmatic understanding: Fine-tuning strategy matters for implicature resolution by llms}.{\BBCQ}
\newblock
\APACjournalVolNumPages{Advances in Neural Information Processing Systems}{36}{}{}.
\PrintBackRefs{\CurrentBib}

\bibitem [\protect \citeauthoryear {%
Sravanthi%
\ \protect \BOthers {.}}{%
Sravanthi%
\ \protect \BOthers {.}}{%
{\protect \APACyear {2024}}%
}]{%
sravanthi-etal-2024-pub}
\APACinsertmetastar {%
sravanthi-etal-2024-pub}%
\begin{APACrefauthors}%
Sravanthi, S.%
, Doshi, M.%
, Tankala, P.%
, Murthy, R.%
, Dabre, R.%
\BCBL {}\ \BBA {} Bhattacharyya, P.%
\end{APACrefauthors}%
\unskip\
\newblock
\APACrefYearMonthDay{2024}{{\APACmonth{08}}}{}.
\newblock
{\BBOQ}\APACrefatitle {{PUB}: A Pragmatics Understanding Benchmark for Assessing {LLM}s{'} Pragmatics Capabilities} {{PUB}: A pragmatics understanding benchmark for assessing {LLM}s{'} pragmatics capabilities}.{\BBCQ}
\newblock
\BIn{} L\BHBI W.~Ku, A.~Martins\BCBL {}\ \BBA {} V.~Srikumar\ (\BEDS), \APACrefbtitle {Findings of the Association for Computational Linguistics: ACL 2024} {Findings of the association for computational linguistics: Acl 2024}\ (\BPGS\ 12075--12097).
\newblock
\APACaddressPublisher{Bangkok, Thailand}{Association for Computational Linguistics}.
\newblock
\begin{APACrefURL} \url{https://aclanthology.org/2024.findings-acl.719} \end{APACrefURL}
\newblock
\begin{APACrefDOI} \doi{10.18653/v1/2024.findings-acl.719} \end{APACrefDOI}
\PrintBackRefs{\CurrentBib}

\bibitem [\protect \citeauthoryear {%
Srivastava%
\ \protect \BOthers {.}}{%
Srivastava%
\ \protect \BOthers {.}}{%
{\protect \APACyear {2022}}%
}]{%
srivastava2023-BIGbench}
\APACinsertmetastar {%
srivastava2023-BIGbench}%
\begin{APACrefauthors}%
Srivastava, A.%
, Rastogi, A.%
, Rao, A.%
, Shoeb, A\BPBI A\BPBI M.%
, Abid, A.%
, Fisch, A.%
\BDBL {}others%
\end{APACrefauthors}%
\unskip\
\newblock
\APACrefYearMonthDay{2022}{}{}.
\newblock
{\BBOQ}\APACrefatitle {Beyond the imitation game: Quantifying and extrapolating the capabilities of language models} {Beyond the imitation game: Quantifying and extrapolating the capabilities of language models}.{\BBCQ}
\newblock
\APACjournalVolNumPages{arXiv preprint arXiv:2206.04615}{}{}{}.
\PrintBackRefs{\CurrentBib}

\bibitem [\protect \citeauthoryear {%
Tong%
, Shutova%
\BCBL {}\ \BBA {} Lewis%
}{%
Tong%
\ \protect \BOthers {.}}{%
{\protect \APACyear {2021}}%
}]{%
tong-etal-2021-recent}
\APACinsertmetastar {%
tong-etal-2021-recent}%
\begin{APACrefauthors}%
Tong, X.%
, Shutova, E.%
\BCBL {}\ \BBA {} Lewis, M.%
\end{APACrefauthors}%
\unskip\
\newblock
\APACrefYearMonthDay{2021}{{\APACmonth{06}}}{}.
\newblock
{\BBOQ}\APACrefatitle {Recent advances in neural metaphor processing: A linguistic, cognitive and social perspective} {Recent advances in neural metaphor processing: A linguistic, cognitive and social perspective}.{\BBCQ}
\newblock
\BIn{} K.~Toutanova\ \BOthers {.}\ (\BEDS), \APACrefbtitle {Proceedings of the 2021 Conference of the North American Chapter of the Association for Computational Linguistics: Human Language Technologies} {Proceedings of the 2021 conference of the north american chapter of the association for computational linguistics: Human language technologies}\ (\BPGS\ 4673--4686).
\newblock
\APACaddressPublisher{Online}{Association for Computational Linguistics}.
\newblock
\begin{APACrefURL} \url{https://aclanthology.org/2021.naacl-main.372/} \end{APACrefURL}
\newblock
\begin{APACrefDOI} \doi{10.18653/v1/2021.naacl-main.372} \end{APACrefDOI}
\PrintBackRefs{\CurrentBib}

\bibitem [\protect \citeauthoryear {%
Wei%
\ \protect \BOthers {.}}{%
Wei%
\ \protect \BOthers {.}}{%
{\protect \APACyear {2022}}%
}]{%
wei2022chain}
\APACinsertmetastar {%
wei2022chain}%
\begin{APACrefauthors}%
Wei, J.%
, Wang, X.%
, Schuurmans, D.%
, Bosma, M.%
, Chi, E.%
, Le, Q.%
\BCBL {}\ \BBA {} Zhou, D.%
\end{APACrefauthors}%
\unskip\
\newblock
\APACrefYearMonthDay{2022}{}{}.
\newblock
{\BBOQ}\APACrefatitle {Chain of thought prompting elicits reasoning in large language models} {Chain of thought prompting elicits reasoning in large language models}.{\BBCQ}
\newblock
\APACjournalVolNumPages{arXiv preprint arXiv:2201.11903}{}{}{}.
\PrintBackRefs{\CurrentBib}

\end{thebibliography}
\clearpage
\section*{Supplementary Materials}
\begin{figure*}[h!]
    \centering
    \includegraphics[width=0.9\textwidth]{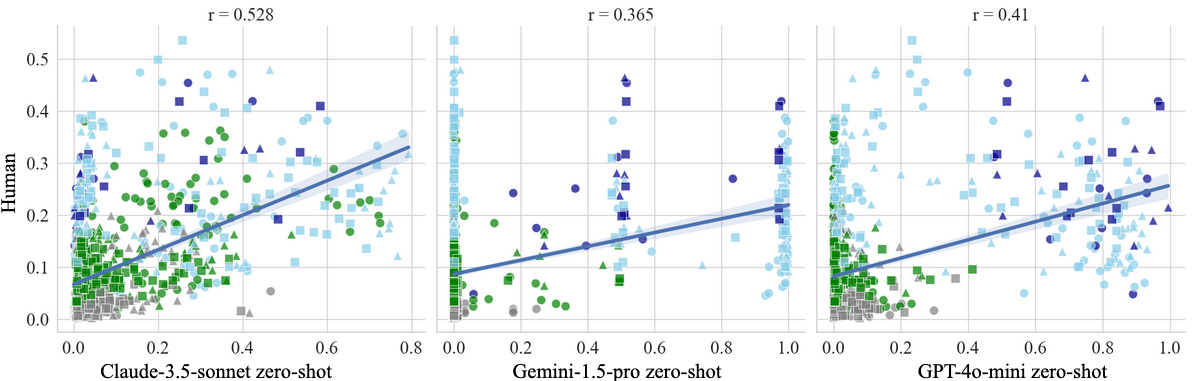} 
    \caption{Probabilities of each pair of $(u, s)$, predicted by different LLMs under \textbf{zero-shot} prompting (with $\tau=1$) (facets, x-axis), plotted against human results (y-axis), coded for each type of interpretation (color) and item (dot shape).}
    \label{fig:expt1}
\end{figure*}

\begin{table}[h]
\centering
\begin{tabular}{llll}
\toprule
 \textbf{LLM} & GPT & Claude & Gemini \\ \midrule
 1-shot price prior CoT& 0.7  & 0.596 & 0.774 \\
 1-shot speaker goals CoT & 0.472 & 0.552 & 0.501 \\
 full LM-based RSA & 0.783  & 0.80 & 0.76 \\
 \bottomrule
\end{tabular}
\caption{Correlation between human predictions of $(s,u)$ probabilities, and results from LLMs under different RSA-inspired approaches. Results for ablations of the CoT prompt are presented where only reasoning about the prior (1-shot price prior CoT) or the speaker communicative goal (1-shot speaker goal CoT) are included. LLM-based RSA refers to results of the full RSA model with both LLM priors and LLM speaker likelihoods. \label{tab:gpt-prompting-comparison}}
\end{table}

\subsection{LLM Zero-Shot performance}
\label{sec:app:zero-shot}
To test whether LLMs arrive at non-literal meanings of
numbers when people do, we closely follow the procedure
and the scenarios presented in \citet{kao2014nonliteral}. To this end, we construct zero-shot prompts to sample LLMs' judgments of probabilities of different true prices $s \in S$, given a speaker's utterance mentioning a price $u \in U$. An example prompt is presented in~\autoref{tab:prompts}.
Results of LLM predictions for all items and all $(u,s)$ pairs are shown against human results in~\autoref{fig:expt1}. 
Under zero-shot prompting, LLMs did not show high correlation with human results, instead showing a tendency towards literal interpretations. 
Furthermore, different models exhibited distinct distributional patterns: GPT-4o-mini tended to assign inflated probabilities to individual utterance-meaning pairs, while Gemini-1.5-pro generally exhibited a bimodal distribution of ratings at the ends of the scale.

\subsection{Guiding LLM Interpretation with the RSA model}
\label{sec:prompting}

To test if the computational steps formalized by the RSA model
can be used to guide LLMs’ interpretation of hyperbole and pragmatic halo in a more human-like way, we compare two possible approaches to integrating the RSA model with LLMs. 

First, we construct a one-shot chain-of-thought (CoT) prompt that verbally describes critical components within the RSA model: reasoning about possible speaker goals and priors of prices for an example every-day item (a toaster).
The full prompt is shown in~\autoref{prompt:rsa}.
\begin{figure}[htpb]
\centering
\begin{tcolorbox}[
width=1\linewidth,
title={One-Shot RSA Prompt}]
\fontsize{5pt}{5pt}\selectfont
\ttfamily
\begin{lstlisting}[language={}]
EXAMPLE:
Anne bought a new toaster. A friend asked her, "Was it expensive?" Anne said, "It cost \$1000."
Please provide the probability that Anne thinks that the toaster is expensive.
Let's think step by step and consider Anne's goals. To answer her friend's question, Anne might want to tell her friend the price, so that her friend can judge whether the toaster is expensive or not. 
She could have the goal to communicate the exact price, or to communicate her attitude about the price or both.
Anne said "\$1000", but given general world knowledge, it is unlikely that a toaster costs literally \$1000. Therefore, it is unlikely that Anne wants to communicate the exact price. A toaster that costs \$1000 would be considered expensive, which would be upsetting. Therefore, it is more likely that Anne wants to communicate that she is upset and felt that the toaster was too expensive, using a hyperbole to talk about the price.
Therefore, it is likely that Anne thinks that the toaster is expensive. The answer is: 0.9
A: 0.9
\end{lstlisting}
\end{tcolorbox}
\caption{\textbf{One-Shot RSA Prompt}
The system prompt and one-shot chain-of-thought prompt for teaching a model to simulate an RSA-model.}
\label{prompt:rsa}
\end{figure}

Results reported in~\autoref{tab:cor_table} (1-shot RSA CoT) indicate that the RSA-couched prompting effectively helped to guide LLMs towards more human-like interpretation, improving the correlation between LLM predictions and human data from \citet{kao2014nonliteral}.

To critically assess the robustness of the prompting and which aspects of the prompt really drive the performance improvements, we ablate parts of the prompt corresponding to different computational components of the RSA model.
Specifically, we construct a speaker-goals prompt which only exemplifies reasoning about different speaker goals (see~\autoref{prompt:rsa-qud}), and a priors prompt which only exemplifies reasoning about price priors (see~\autoref{prompt:rsa-priors}).
\begin{figure}[htpb]
\centering
\begin{tcolorbox}[
width=1\linewidth,
title={Ablated QUD-only One-Shot Prompt}]
\fontsize{5pt}{5pt}\selectfont
\ttfamily
\begin{lstlisting}[language={}]
In each scenario, two friends are talking about the price of an item.
Please read the scenarios carefully and provide the probability that the item has the desribed price.
Provide the estimates on a continuous scale between 0 and 1, where 0 stands for "impossible" and 1 stands for "extremely likely".
Write ONLY your final answer as 'A:<rating>'.

EXAMPLE:
Anne bought a new toaster. A friend asked her, "Was it expensive?" Anne said, "It cost $1000."
Please provide the probability that the toaster cost $50.
Let's think step by step and consider the possible communicative goals of Anne.
Anne might want to communicate about the price, about her attitude towards the price, or both.
For communicating the price, she would choose to be precise, ignoring other possible meaning dimesnions. For communicating her attitude, she would choose a an expression that signal attitude, where other possible dimensions like being precise don't matter. For communicating both, she might choose an utterance that trades off both goals. 
Thr utterance seems to fit the goals attitude communication and both. Therefore, the answer is: 0.75
A: 0.75

YOUR TURN:

\end{lstlisting}
\end{tcolorbox}
\caption{\textbf{Ablated QUD-only One-Shot Prompt}
The system prompt and one-shot chain-of-thought prompt for teaching a model to reason about the communicative goals, as suggested by the RSA-model.}
\label{prompt:rsa-qud}
\end{figure}
\begin{figure}[htpb]
\centering
\begin{tcolorbox}[
width=1\linewidth,
title={Ablated Priors-only One-Shot Prompt}]
\fontsize{5pt}{5pt}\selectfont
\ttfamily
\begin{lstlisting}[language={}]
In each scenario, two friends are talking about the price of an item.
Please read the scenarios carefully and provide the probability that the item has the desribed price.
Provide the estimates on a continuous scale between 0 and 1, where 0 stands for "impossible" and 1 stands for "extremely likely".
Write ONLY your final answer as 'A:<rating>'.

EXAMPLE:
Anne bought a new toaster. A friend asked her, "Was it expensive?" Anne said, "It cost $1000."
Please provide the probability that the toaster cost $50.
Let's think step by step and consider the prior probability of toaster prices.
Given general world knowledge, it is unlikely that a toaster costs literally $1000. Rather, a price around $50 would be considered a normal price for a toaster. Therefore, a toaster that costs $1000 would be considered expensive. 
Since Anne stated an unlikely price for the toaster, it is likely that the true price of the toaster was not what would normally be expected a priori. Therefore, the answer is: 0.75
A: 0.75

YOUR TURN:


\end{lstlisting}
\end{tcolorbox}
\caption{\textbf{Ablated Priors-only One-Shot Prompt}
The system prompt and one-shot chain-of-thought prompt for teaching a model to reason about the priors of prices of an item, as suggested by the RSA-model.}
\label{prompt:rsa-priors}
\end{figure}
Results of these ablations as measured by the correlation with human data are presented in~\autoref{tab:gpt-prompting-comparison}.
Compared to the full one-shot CoT prompt, the speaker goals only prompt led to lower correlation between LLM and human data for all LLMs.
The priors only prompt, on the other hand, increased the correlation between LLM and human results more strongly than the full one-shot CoT prompt (see~Table~\ref{tab:cor_table} in main text). 
These ablation results suggest
that LLM performance can be supported through RSA model
inspired prompting, but the prompt components required for substantially increasing LLM performance may not necessarily have to fully replicate all the computational components
needed for explaining human performance.

\subsection{LM-RSA Simulations}
Second, we used the RSA model to quantify the LLMs' internal consistency between its own predicted priors and zero-shot prompting based predictions.
To this end, we used the priors of prices for different items and for priors for affect, given a price, predicted by LLMs, elicited in Experiment~3.
We then fit the RSA model proposed by \citet{kao2014nonliteral} using the priors from each LLM, resulting in the \textit{LM-RSA model}.
The RSA model includes two hyperparameters that were fit to human behavioral data. 
To adjust for biases against using the extreme probability ratings for the $(u,s)$ pairs, a power-law transformation was performed: we
multiplied the predicted probability for each $(u,s)$ pair by a free parameter $\lambda$, and renormalized the probabilities to sum up to 1 for each utterance $u$. 
The $\lambda$ was jointly fit with the model’s cost ratio $C$. 
$C(u) = 1$ was used when $u$ was a round number (divisible by 10) and the cost for sharp utterances was fit to human data.
We tune the cost and $\lambda$ hyperparameters individually for each LM-RSA. 
The optimal $\lambda$ was chosen via search over $[0, 1)$ with steps of 0.01. The optimal $C$ was chosen via search over $[1, 4)$ with steps of 
0.1. The best hyperparameters which were used to produce results reported in the main text are shown in~\autoref{tab:rsa-hyperparams}.
\begin{table}[h]
\centering
\begin{tabular}{lll}
\toprule
                   & $\lambda$ & $C$ \\ 
\midrule
Human priors       & 0.44   & 1.2  \\ 
GPT-4o-mini priors & 0.41   & 1.4   \\ 
Claude-3.5-sonnet priors & 0.39   & 2.0   \\ 
Gemini-1.5-pro priors & 0.38 & 1.1 \\ 
\bottomrule
\end{tabular}
\caption{Best hyperparameters of the RSA model, fit separately for each LLM-RSA model. \label{tab:rsa-hyperparams}}
\end{table}

\subsection{Eliciting LLMs' Utterance Likelihoods}
In Experiment~3, we focused on assessing key aspects that might be the root cause of LLMs' deficiencies in non-lieral number interpretation, informed by the RSA model: 
(1) the price priors that LLMs assign to for each item, (2) the priors of affective responses conditional on item prices, and (3) the LLMs' representations of speaker likelihoods of uttering different prices $u$, given different true price states $s$ and speaker goals. 
The low correlation of LM-RSA, zero-shot LLM and human predictions (Experiment 3) revealed that the challenge in achieving human-like pragmatic interpretation of hyperbole and halo was not due to lack in the models’ price and affect priors.
Therefore, we investigated whether the LLM might lack the third computational component that constitutes the pragmatic interpreter in the RSA model: the pragmatic speaker $S_1$.  
Specifically, we tested whether the LLMs captured the likelihoods of different utterances $u \in U$ conveying the speaker's communicative goal and intended meaning about an item with sufficient accuracy. 

We collected likelihoods of uttering a price $u \in U$, given different states and goals of the speaker based on \textit{zero-shot prompting}, as shown in~\autoref{prompt:affect}.
\begin{figure}[htpb]
\centering
\begin{tcolorbox}[
width=1\linewidth,
title={Pragmatic Speaker Prompt}]
\fontsize{5pt}{5pt}\selectfont
\ttfamily
\begin{lstlisting}[language={}]
In each scenario, two friends are talking about the price of an item.
Please read the scenarios carefully and provide the probability that the speaker would say the following utterance, given their communicative goal and the true price of the item.
Provide the estimates on a continuous scale between 0 and 1, where 0 stands for "impossible" and 1 stands for "extremely likely".
Write ONLY your final answer as 'A:rating'.
Bob bought a laptop. The laptop cost \$100. A friend asked Bob if the laptop was expensive. 
Bob wants to communicate both their attitude towards the price of the laptop they bought and the price of the laptop. 
Bob wants to precisely communicate the price of the laptop they bought.
Bob thinks the laptop is too expensive.
How likely is it that Bob will say: 'The laptop cost \$1000.'?
\end{lstlisting}
\end{tcolorbox}
\caption{\textbf{Pragmatic Speaker Prompt}
The system prompt and one-shot prompt for asking a model to predict the probability of a specific utterance, when communicating \textit{both affect and the exact state}. The sentence about the affect and the preciseness of the price are removed, if the communicative goal only contains one dimension, respectively.}
\label{prompt:affect}
\end{figure}
We used these raw LLM predictions to calculate the speaker likelihood $P_{LLM}(u \mid s, a, g)$, where $g \in G$ is the speaker goal (communicating exact or round price, and communicating the price, the affect, or both), $a$ represented affect and could take on the values 0 or 1, resulting in 12 condition, some of which are identical.
We calculated the final $P_{LLM}(u \mid s, a, g)$ by aggregating over the raw LLM predictions over the irrelevant meaning dimensions as suggested in the RSA model (repeated from main text): 
$$S_1 (u \mid s, a, g) \propto \sum_{s', a'} \delta_{g(s, a) = g(s', a')}L_0(s', a' \mid u) \cdot e^{-c(u)}$$
That is, for instance, when the goal $g$ is to convey affect only, the scores are aggregated and renormalized across the values of $S$.  

We investigated the accuracy of the utterance likelihood representations of the LLMs by fitting a \textit{full LM RSA model}.
The $P_{LLM}$ was used together with LLM priors elicited from the same LLM in Experiment 3 to fit this RSA model (via enumeration) and predict posteriors of different meanings, given utterances. 
 
We report the correlation of the full LM RSA posteriors of all $(u, s)$ with human data in \autoref{tab:gpt-prompting-comparison} (``full LM-based RSA''). 
We found that the full LM-RSA model showed notably higher correlation with human data than the zero-shot predictions of the same LLM models under zero-shot prompting (see~\autoref{tab:cor_table} in the main text).
The correlations are also higher than for the one-shot RSA CoT based results (\autoref{tab:cor_table}), suggesting that more explicit RSA-based task decomposition might guide LLMs towards more human-like interpretation in a more stringent way.
In other words, the computational components of the RSA model provide a structure which allows to build a fully LLM-based system that rationalizes an observed utterance in a more human-like way, based  only on assuming the space of possible speaker goals.
At the same time, the correlations of the full LM-RSA models are lower than for the LM-RSA model (based only on LLM priors), suggesting that LLMs' utterance likelihoods might be less human-like than LLMs' priors. 
Additioanlly, the correlations of full LM-RSA models are ordinally the same as the zero-shot predictions (from Gemini to Claude). 
Taken together, these comparisons suggest that the lack in representing human-like utterance likelihoods might contribute to the LLMs' interpretation difficulties in a zero-shot setting

These detailed analyses open up interesting avenues for investigating why LLM utterance likelihoods differ compared to humans, and whether, e.g., particular training or dataset aspects lead to the discrepancy between humans and LLMs.
We turn to some explorations in the next section.

\subsection{Free Generation of Prices}
Based on the main results, we report another exploratory investigation as to \textit{why} the LLMs may have deficient performance on hyperbole and halo, despite correctly representing component information suggested by then RSA model. 

One potential reason could be that the materials from \citet{kao2014nonliteral} used in Experiment 1 as the state space $S$ were out-of-distribution for the LLMs.
Focusing on GPT-4o-mini, we ran an exploratory free generation study, identifying which prices the LLM would generate under $\tau=1$ when prompted to complete the speaker's utterance about the price of an item, given the speaker's goal (e.g., to convey the state, the affect, or both; being precise or imprecise about the price). 
We then qualitatively assessed whether the produced numbers differed from $U$ = \{50 + k, 500 + k, 1000 + k, 5000 + k, 10000 + k\}, with $k \in \{0, 1, 2, 3\}$.
Additionally, we compared whether LLMs freely generate higher prices when prompted to convey affect (i.e., that the item was expensive), than when prompted without affect . The full prompt can be found in~\autoref{prompt:free-generation}. 
\begin{figure}[htpb]
\centering
\begin{tcolorbox}[
width=1\linewidth,
title={Free Generation Prompt}]
\fontsize{5pt}{5pt}\selectfont
\ttfamily
\begin{lstlisting}[language={}]
In each scenario, two friends are talking about the price of an item.
Please read the scenarios carefully and complete the speaker's utterance with your best guess, given their communicative goal and the true price of the item.
Write ONLY your numerical completion for the utterance as 'A:<completion>'.
\end{lstlisting}
\end{tcolorbox}
\caption{\textbf{Free Generation Prompt}
The system prompt for asking a model to generate the likely utterance mentioning a price, when communicating a particular goal.}
\label{prompt:free-generation}
\end{figure}

Based on one exploratory simulation for all items and prices, we found that LLM produced different prices than in $U$. For instance, the predicted utterances for the electric kettle ranged from \$30 to \$150. The LLM produced higher prices when prompted to communicate that the speaker think the item is expensive (under the goal to communicate affect or both meaning dimensions), than when the affect was not present (mean predicted prices across items: \$342~vs.~\$590). These results suggests that the LLM are, in principle, able to generate hyperbolic utterances, but may be sensitive to the exact price numbers when prompted to interpret utterances or assess the likelihood of particular utterances.

\end{document}